\documentclass[letterpaper, 10 pt, conference]{ieeeconf}  % Comment this line out if you need a4paper
\PassOptionsToPackage{table}{xcolor}
\IEEEoverridecommandlockouts                        % This command is only needed if 
\usepackage[T1]{fontenc}
\usepackage{amsmath,amsfonts}
\usepackage{algorithm}
\usepackage{graphicx}
\usepackage{float}
\usepackage{lipsum}
\usepackage{algpseudocode}
\usepackage{pgfplots}
\pgfplotsset{compat=1.17}
\usepackage{array} % Add this in your preamble
\usepackage{multirow} % Add this line to the preamble
\usepackage{threeparttable} % For tablenotes environment
\usepackage{makecell}
\usepackage{siunitx}
\usepackage{tabularx}
\usepackage[table]{xcolor}
\usepackage{flushend}
\usepackage{dblfloatfix}    % or stfloats
\usepackage[font=footnotesize, labelfont=footnotesize]{caption}
\usepackage{dirtytalk}
\usepackage{subcaption}

\definecolor{rowlabelcolor}{RGB}{240, 246, 255}% Very light, airy blue
\newcolumntype{P}[1]{>{\centering\arraybackslash}p{#1}} % Defines a centered version of p{}
\usepackage[hang,flushmargin]{footmisc}
\usepackage{booktabs}    % For \midrule and other table enhancements
\usepackage{adjustbox}   % For trimming images
\usepackage{microtype}
\usepackage{booktabs}
\usepackage{cite}

\makeatletter
\let\NAT@parse\undefined
\makeatother
\usepackage[breaklinks,colorlinks]{hyperref}

\renewcommand{\baselinestretch}{0.980}
\begin{document}

\title{\LARGE \bf
Contour Errors: Ego-Centric Matching for\\
3D Multi-Object Tracking Performance Evaluation}

\author{Sharang Kaul$^{1,2}$, Simon Bultmann$^{2}$, Mario Berk$^{1}$ and Abhinav Valada$^{2}$
 \thanks{$^1$CARIAD SE, Germany}%
 \thanks{$^2$Department of Computer Science, University of Freiburg, Germany}%
}

\maketitle
\thispagestyle{empty}
\pagestyle{empty}

%%%%%%%%%%%%%%%%%%%%%%%%%%%%%%%%%%%%%%%%%%%%%%%%%%%%%%%%%%%%%%%%%%%%%%%%%%%%%%%%
\begin{abstract}
Open-loop performance evaluation of 3D multi-object tracking in autonomous driving requires matching criteria that effectively penalize translational, shape, and orientation errors from the ego-vehicle perspective. The prevailing criteria for determining true positives are Intersection over Union (IoU) and Centre-Point Distances (CPD). When IoU is extended from the 2D image plane to 3D volumetric overlap, it often falls below its acceptance threshold even with minor yaw misalignments, whereas CPD disregards orientation entirely. To address this limitation, we propose Contour Errors (CE) as an ego-centric criterion that employs Hausdorff-type reasoning to sparse bounding-box corner geometry by selecting the $k$-nearest ego-centric corners. This method provides a graded orientation sensitivity between the extremes of IoU, which over-penalizes, and CPD, which is orientation-blind. We evaluate Contour Errors against six baselines using the HOTA evaluation protocol on the nuScenes dataset, conditioned on proximity, yaw error, and a confidence threshold. At the standard IoU vehicle threshold, 47\% of car and 75\% of pedestrian CE-valid matches are rejected by IoU despite close contour proximity, while fewer than 0.1\% of IoU-valid matches fail CE. These results establish the ego-centric matching criterion as a primary factor for improving open-loop perception evaluation in safety-critical autonomous driving.
\end{abstract}

%%%%%%%%%%%%%%%%%%%%%%%%%%%%%%%%%%%%%%%%%%%%%%%%%%%%%%%%%%%%%%%%%%%%%%%%%%%%%%%%

\section{Introduction}

In autonomous driving, multi-object tracking (MOT) evaluation relies on a matching criterion that, for each frame, assigns ground truth and predicted object pairs to true positives (TPs), false negatives (FNs), and false positives (FPs)~\cite{lang2022robust, lang2023self}. The two main criteria for object matching are Intersection over Union (IoU)~\cite{padilla2020survey}, which employs a threshold of $\geq 0.7$ for vehicles in the KITTI~\cite{geiger2012we} and Waymo~\cite{sun2020scalability} datasets, and Center Point Distance (CPD), which uses a threshold of $\leq 2$\,\si{\meter}~\cite{bernardin2008evaluating}, as adopted in nuScenes~\cite{caesar2020nuscenes} and Argoverse~2~\cite{wilson2023argoverse}. Both criteria are object-centric, evaluating each detection within its local coordinate frame and disregarding the ego vehicle's perspective~\cite{leal2017tracking}. In three-dimensional space, this approach leads to two complementary failure modes~\cite{buchner20223d}. IoU often rejects functionally adequate matches; even minor yaw misalignments, particularly around the vertical $z$-axis, can substantially reduce volumetric intersection. nuScenes adopted CPD as IoU is ineffective for small-footprint objects with minor translation errors~\cite{caesar2020nuscenes}. In contrast, CPD is entirely orientation-insensitive and accepts predictions with any heading error.

\begin{figure}[t!] 
\centering 
\includegraphics[width=\columnwidth]{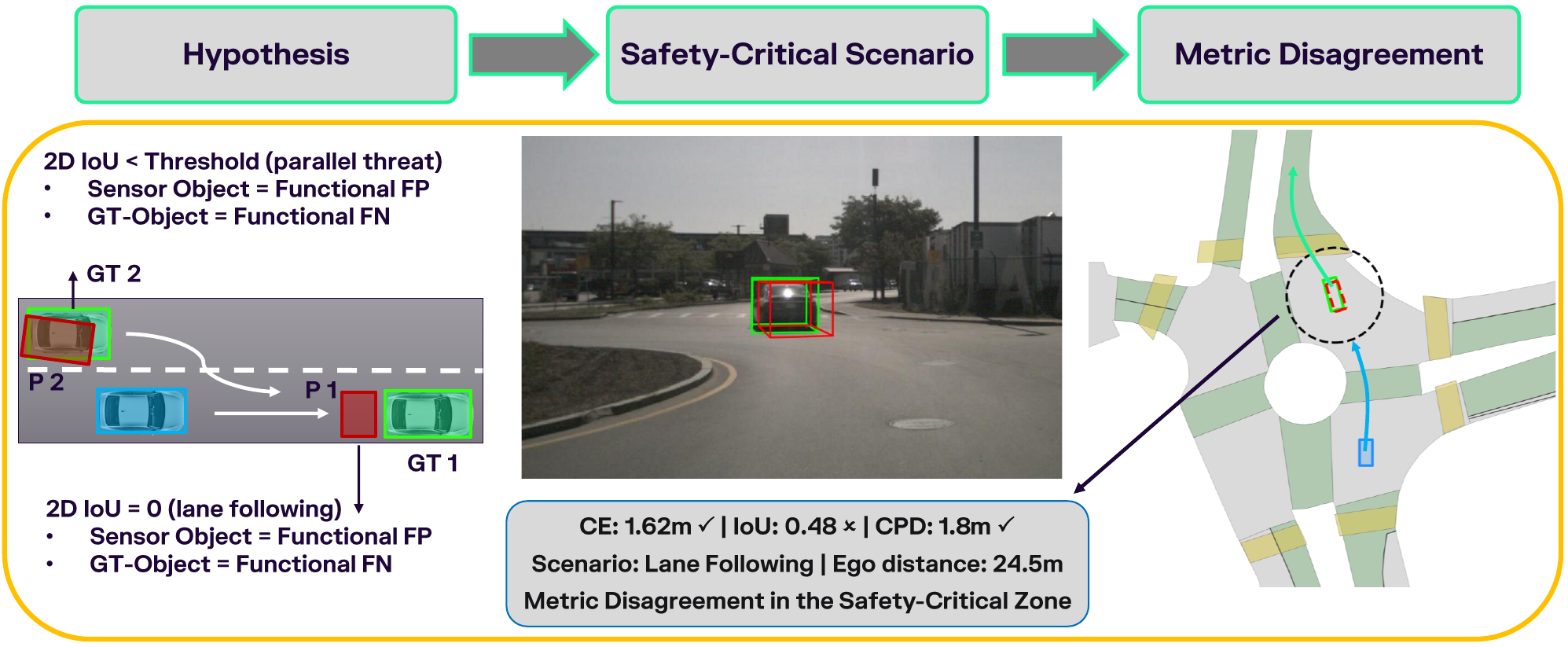} \vspace{1pt} 
\includegraphics[width=\columnwidth]{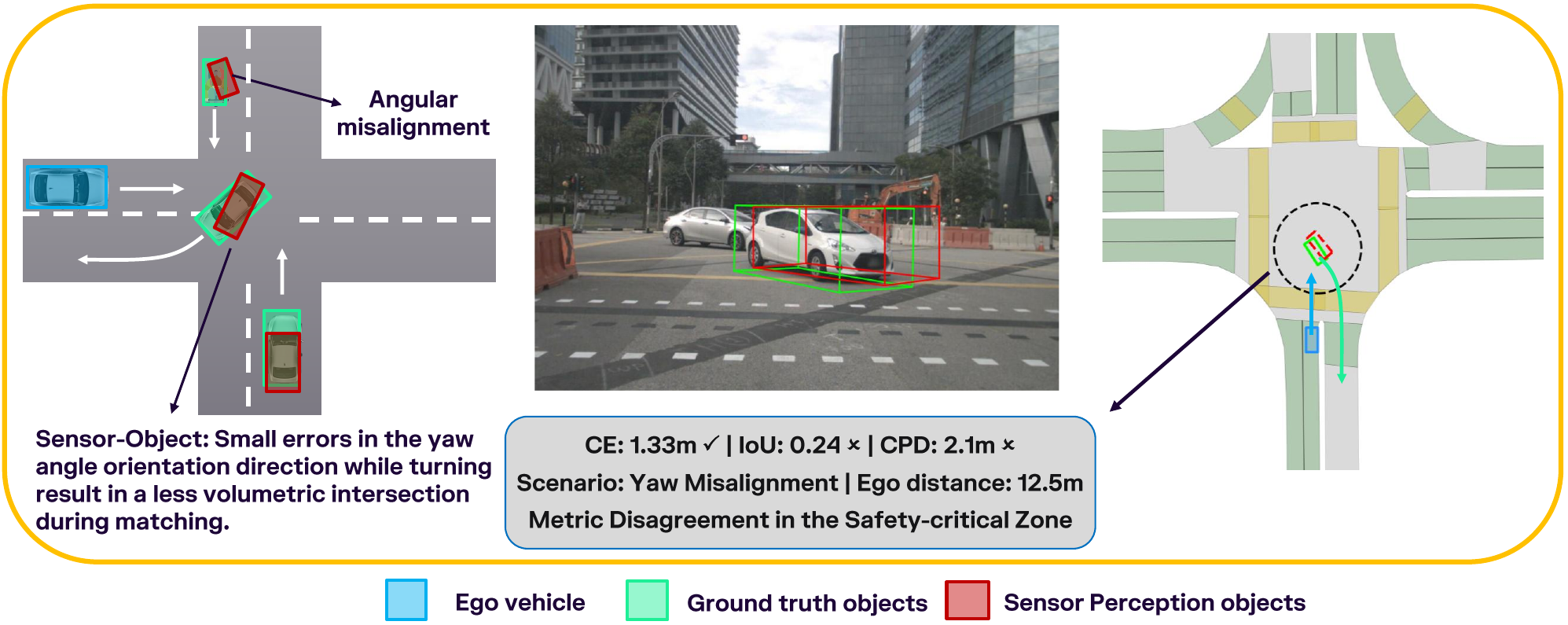}
\caption{Motivation for ego-centric evaluation on the nuScenes dataset. Each row represents: the scenario hypothesis (left), the camera view with green ground-truth labels and red predictions  (middle), and the BEV map (right). The top and bottom rows represent two hypothesis: (a)~\emph{Partial visibility} and (b)~\emph{Yaw misalignment}, respectively, as explained in Sec.~\ref{sec:hypothesis_2}}.
\label{fig:egocentric_view}
\vspace{-0.9cm}
\end{figure}

This study investigates open-loop perception assessment for safety-critical autonomous driving, focusing on situations in which a planner is not yet available, for example in the early development stages. In these situations, the matching criterion determines the measured detection quality, and a higher TP count does not necessarily indicate superior evaluation. Predictions that exhibit only small yaw deviations should be considered TPs, while those with substantial heading errors should be excluded. IoU penalizes both situations too strongly, whereas CPD is completely insensitive to yaw errors. The CE metric provides a graduated, orientation-sensitive response between these extremes, as illustrated in Fig.~\ref{fig:egocentric_view}. Within an ego-centric reference frame, such as that of a moving vehicle, object positions and orientations change continuously relative to the ego agent (see Fig.~\ref{fig:egocentric_view}). Intersection-based metrics often assign low IoU values even when bounding boxes are nearly correct but exhibit minor shifts or rotations. Similarly, CPD fails to capture yaw misalignments, which are critical for downstream tasks such as object tracking and motion forecasting. These geometric inconsistencies between target objects and the ego vehicle impede meaningful error analysis, particularly in scenarios involving partial or non-overlapping detections within complex three-dimensional scenes.

To address these limitations, this work introduces 2D and 3D Contour Errors (CEs) for ego-centric evaluation, a novel matching criterion that resolves inconsistencies in object matching from an ego-centric perspective. The analysis centers on autonomous vehicle environments, where precise object matching is essential for reliable perception and decision-making. Two scenario-based hypothesis are formulated, and CE is evaluated against prevailing multiple matching criteria.

\textit{Partial Visibility}: \label{sec:hypothesis_1}
In the lane-following scenario depicted in Fig.~\ref{fig:egocentric_view} (top), only the rear face of the lead vehicle is visible. IoU penalizes the match because the predicted bounding box does not overlap the entire ground truth (GT) extent, resulting in a false negative despite geometric proximity. Contour Error (CE) evaluates only the $k$ nearest ego-centric corners, maintaining robustness to unobserved regions of the bounding box.

\textit{Yaw Misalignment at Intersections}: \label{sec:hypothesis_2}
At an urban intersection, as shown in Fig.~\ref{fig:egocentric_view} (bottom), vehicles approach with significant changes in heading. IoU falls below its threshold when the predicted bounding box is rotated relative to the ground truth, even if both boxes are spatially proximate. CPD, which is orientation-insensitive, accepts misoriented matches and thus fails to capture safety-critical heading errors. Contour Error (CE) quantifies the discrepancy in the ego-facing contour, penalizing substantial yaw offsets while accommodating uncertainty in depth.

The primary contributions of this work are as follows: (i) the introduction of Contour Errors (CEs), which adapt Hausdorff-type reasoning from a geometric discrepancy measure into an ego-centric, evaluation-time matching criterion for sparse bounding-box corners using $k$-nearest ego-centric corner selection; to the best of our knowledge, this formulation has not previously been applied as a matching criterion in multi-object tracking (MOT) evaluation, (ii) a comprehensive evaluation within the Higher Order Tracking Accuracy (HOTA) protocol on the nuScenes dataset, demonstrating that the choice of matching criterion primarily influences frame-level detection quality (DetA), while association accuracy (AssA) remains largely unchanged, and (iii) proximity-, yaw-, and confidence-conditioned analysis showing that IoU excessively penalizes minor orientation deviations, CPD is completely insensitive to heading errors, and CE provides a principled intermediate approach with adjustable orientation sensitivity.

%%%%%%%%%%%%%%%%%%%%%%%%%%%%%%%%%%%%%%%%%%%%%%%%%%%%%%%%%%%%%%%%%%%%%%%%%%%%%%%%
\section{Related Work}

% {\parskip=0pt
% \noindent\textit{Object Detection Metrics}: \cut{Metrics for the evaluation of object detection have significantly evolved in autonomous driving. These metrics are essential for the robust evaluation of 3D perception tasks. Traditional metrics such as Precision, Recall, and Average Precision (AP) remain state-of-the-art for 2D evaluation, often utilizing the IoU thresholds to determine the detection quality~\cite{nallapareddy2023evcenternet, lang2022hyperbolic}. However, in 3D object detection, specialized metrics are required to capture the complexity of spatial orientation, depth, and velocity. Metrics such as mean Average Precision (mAP), widely used in datasets such as KITTI~\cite{geiger2012we} and Waymo Open Dataset~\cite{sun2020scalability}, extend IoU-based evaluation to the 3D domain. The nuScenes Detection Score (NDS)~\cite{caesar2020nuscenes} further enhances mAP by integrating attributes such as orientation, velocity, and object attributes, reflecting the dynamic nature of real-world driving.}}

{\parskip=2pt
\noindent\textit{Overlap-Based Matching Metrics}: Recently proposed metrics such as \cite{ravi2023addressing, zhang2023shape} aim to address the limitations of IoU in 3D object detection. Notably, Generalized IoU (GIoU)~\cite{8953982} augments IoU with a penalty based on the smallest enclosing box, while Distance-IoU (DIoU) and Complete-IoU (CIoU)~\cite{Zheng2019DistanceIoULF} incorporate CPD and aspect-ratio terms, respectively. These variants were designed primarily as differentiable \emph{training losses}. Their utility as \emph{evaluation-time matching criteria} in 3D MOT has received less attention. Additionally, \cite{zhang2023shape} focuses on the inherent properties of bounding boxes, such as shape and scale, and was proposed to address shortcomings in geometric relationships that conventional IoU-based approaches ignore. These advancements underscore the growing focus on metrics that align with the safety-critical goals of autonomous systems.}

{\parskip=2pt
\noindent\textit{Distance-Based Matching Metrics}: Beyond overlap-based measures, distance metrics offer an alternative criterion for evaluating spatial alignment, particularly in scenarios with low overlap or for non-axis-aligned geometries. The Hausdorff distance (HD)~\cite{huttenlocher2002comparing} and its variants, such as the modified Hausdorff distance~\cite{dubuisson1994modified}, measure the maximum of the minimum distances (or, in the modified variant, the average) between two sets of points, providing a rigorous measure of shape similarity that is less sensitive to volumetric discrepancies than IoU. These metrics have been widely adopted in image segmentation, medical imaging, and point cloud registration for their robustness. More recently, the Chamfer distance~\cite{wu2021balanced} has emerged as a popular differentiable alternative for matching unordered point sets, often used in 3D reconstruction and shape completion tasks~\cite{park2019deepsdf}. While powerful for measuring pure geometric fit, a key limitation of these general-purpose distance metrics in the context of ego-centric perception is their insensitivity to the ego perspective. They do not inherently prioritize errors based on their potential impact on the ego agent's safety or decision-making process.}

{\parskip=2pt
\noindent\textit{Multi-Object Tracking Association}: Recent MOT systems improve online association through center-based feature extraction and occlusion handling~\cite{Cao2023AMT}, multi-level association with intelligent filtering~\cite{Liu20263DMT}, and multi-modal cascade association~\cite{10751781}. These approaches improve the object tracker's association process, whereas CE addresses the complementary evaluation stage question of which ground-truth-prediction pairs should be accepted as matches from an ego-centric perspective in each frame.}

{\parskip=2pt
\noindent\textit{Ego-Centric Metrics}: 
Metrics that incorporate or indirectly address the dynamics and relative geometry of the ego agent can include motion prediction, collision avoidance, and risk assessment within the tracking framework~\cite{rasouli2024novel}. These metrics are not discrete but are integrated or derived within the evaluation frameworks~\cite{badithela2022evaluation}. Ceccarelli~\textit{et~al.}~\cite{ceccarelli2024object} extract knowledge based on object relevance to improve the task of planning the future trajectory of the ego agent. Liao~\textit{et~al.}~\cite{liao2024ec} develop a weighted mechanism to assign a higher score to the predicted box whose groundtruth is close to the ego vehicle. Other metrics quantify the impact of object detection on the ego agent from the planner's perspective by leveraging the dynamic attributes of detections in real-world driving tasks~\cite{philion2020learning}. Unlike EC-IoU and planner-aware metrics, which weight or reweight detections by ego relevance or downstream impact, CE geometrically redefines the evaluation-time matching gate using the ego-facing contour.}

% A related but distinct line of work defines safety-critical metrics that evaluate perception from the perspective of downstream vehicle control rather than geometric matching itself~\cite{hsuan2022usc, ceccarelli2022safety}. Ivanovic~\textit{et~al.}~\cite{ivanovic2022injecting} propose task-aware metrics that assess whether detections are relevant for vehicle control in safety-critical scenarios. While these approaches do not directly address the matching problem that our work focuses on, they highlight the broader motivation: matching criteria should reflect the functional relevance of detection errors, not merely geometric overlap.

% However, conventional IoU-based and CPD-based matching metrics fail to capture these ego-centric constraints, often leading to suboptimal tracking evaluations from an autonomous driving functional perspective. Our proposed Contour Error matching criteria addresses this gap by integrating the geometric properties of target objects within the ego frame, thereby enabling more reliable and stable object associations in tracking-by-detection frameworks.

%%%%%%%%%%%%%%%%%%%%%%%%%%%%%%%%%%%%%%%%%%%%%%%%%%%%%%%%%%%%%%%%%%%%%%%%%%%%%%%%
\section{Contour Errors - An Ego-centric Measure}

\begin{figure*}[t]
    \centering
    \includegraphics[width=0.95\linewidth]{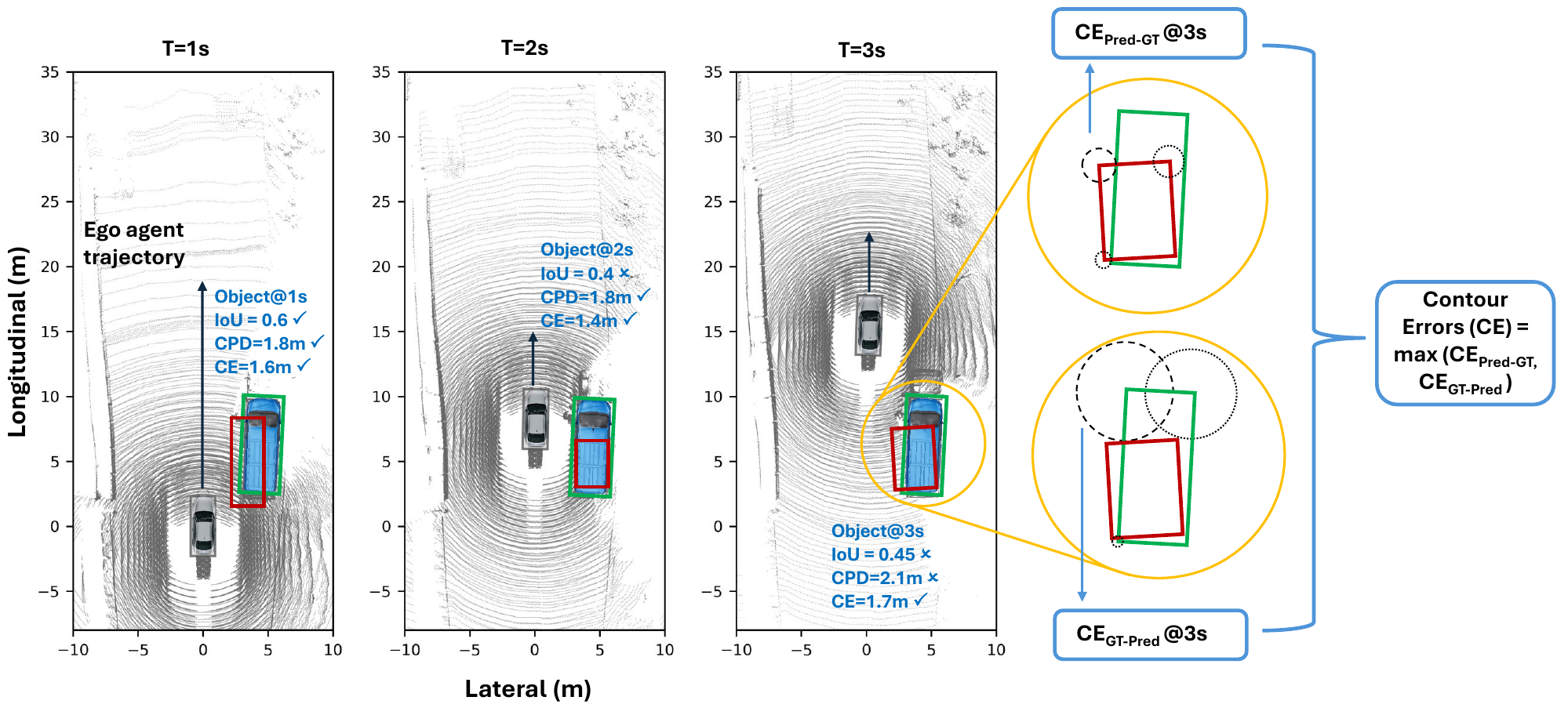}
    \caption{%
    Computation of the ego-centric Contour Error (CE) metric for matching predictions to the ground truth in 2D BEV space. CE is computed \emph{per frame}. The three time steps shown here at \SI{1}{\second}, \SI{2}{\second}, and \SI{3}{\second} illustrate how different metrics evolve, and are not a multi-frame computation. The proposed CE is computed symmetrically for each timestamp. For simplicity, we have shown the calculations for the 3rd timestamp as: (1) the maximum distance from the ground truth (GT) contour in green to the prediction in red ($\text{CE}_\text{GT-Pred}$), and (2) vice-versa from the prediction to the GT ($\text{CE}_\text{Pred-GT}$). The final metric is $\max(\text{CE}_\text{GT-Pred},\, \text{CE}_\text{Pred-GT})$, ensuring consistency under occlusion and perspective change.}
    \label{fig:contour_errors}
%    \vspace{-0.4cm}
\end{figure*}

We introduce contour errors (CEs) as an ego-centric matching criterion that evaluates the proximity of bounding-box surfaces from the ego vehicle's perspective. In the ego-centric version used here, the $k$ nearest corners to the ego centre are selected. The contour error \( E(G_i, P_j) \) between a ground truth object \( G_i \) and a predicted object \( P_j \) is given by

{\small
\begin{align}
E(G_i,P_j)=
\max\!\left( \max_{p\in P'_j}\min_{x\in X_i}\|p-x\|,\,
              \max_{g\in G'_i}\min_{y\in Y_j}\|g-y\|\right)
\end{align}}
where \( G_i' \subseteq G_i \) and \( P_j' \subseteq P_j \) are the $k$ corners of the ground truth and predicted bounding boxes closest to the ego vehicle center. \( X_i \) and \( Y_j \) denote the surfaces of the GT and predicted boxes, respectively. \( \|\cdot\| \) is the Euclidean distance.  A match is established if $E(G_i, P_j) \leq \tau_E$, where \( \tau_E \) is a threshold that defines an acceptable level of shape similarity. For each selected corner, the point-to-surface distance is the Euclidean distance to the nearest face of the opposing box, computed in closed form in $\mathcal{O}(1)$ time per corner.

\textit{Relationship to Hausdorff distance (HD)}: Structurally, CE is a \say{directed HD restricted to the $k$-nearest ego-centric corners}.  The classical symmetric HD between two point sets $A$ and $B$ is $d_H(A,B) = \max\bigl(\max_{a\in A}\min_{b\in B}\|a-b\|,\, \max_{b\in B}\min_{a\in A}\|b-a\|\bigr)$.  CE replaces the full corner sets $G_i,P_j$ with their ego-centric subsets $G_i',P_j'$ of size $k$. When $k$ equals the total number of corners (eight in 3D), CE reduces exactly to the symmetric HD. However, classical HD operates on continuous contours, surfaces, or dense point clouds (e.g., image segmentation, medical imaging, point cloud registration) and, to the best of our knowledge, has not previously been applied as a matching criterion to the discrete corner representation of 3D bounding boxes in MOT evaluation. CE is therefore novel in adapting Hausdorff-type reasoning to sparse box-corner geometry with an ego-centric $k$-nearest restriction that focuses on the corners most relevant to the ego vehicle's safety.

\textit{Bidirectional maximum}: The outer $\max$ in Eq.~(1) ensures that CE penalizes both \emph{under-estimation} representing GT corners far from the prediction surface and \emph{over-estimation} representing prediction corners far from the GT surface.  Using only the GT to predict direction would assign zero error to predictions that are a tiny subset of the GT, thereby missing gross over-detections. Using only Predict to GT would miss size overestimation.  The bidirectional maximum guarantees that neither failure mode is ignored.  Because CE is restricted to the $k$-nearest corners, the far, unobservable side of the bounding box has reduced influence relative to the full Hausdorff, making CE more robust to size-estimation errors on the occluded side while still detecting gross misalignments.

We outline the procedure for computing 2D and 3D Contour Errors in Algo.~\ref{alg:contour_error}. The procedure is identical in both domains. The only difference is the number of selected corners with $k{=}3$ in 2D, $k{=}6$ in 3D, and the dimensionality of the point-to-surface distance computation. Conceptually, one can visualize the distance from each selected corner to the opposing box surface as the radius of a circle (2D) or sphere (3D) centered on that corner and tangent to the nearest point on the opposing box. In a 2D case, as shown in  Fig.~\ref{fig:contour_errors}, we select the three closest corners (e.g., in the lane following scenario, two rear and one frontal corner) of the ground truth box (GT1) to the ego vehicle. $k{=}3$ is the minimum number of corners spanning two adjacent edges required to capture both positional and orientational (yaw) error, as two corners on a single edge cannot detect yaw misalignment.  We then compute the minimum distances from the nearest corners to the predicted bounding box (P1). In 3D, we use spheres centered at the six closest GT vertices (the rear face and one side face in the same scenario) and determine where they contact the predicted box to measure the nearest distances. $k{=}6$ corresponds to the typical two-face visibility geometry when the ego observes a vehicle's rear and one side. Therefore, the selection of corners varies with the domain.

\begin{algorithm}[t]
\footnotesize
\caption{Contour Error-Based Matching}
\label{alg:contour_error}
\textbf{Input:} Ground truth boxes $(G_i)_{i=1}^{n}$, Predicted boxes $(P_j)_{j=1}^{m}$, Threshold $\tau_E$, Dimension $\text{dim} \in \{2, 3\}$\\
\textbf{Output:} Contour distance matrix $D = [d_{ij}]_{n \times m}$

\begin{algorithmic}[1]

\Procedure{Calculate\_Contour\_Error}{\(G_i\), \(P_j\), \text{dim}}
    \State \textbf{1. Select the three and six closest corners to the ego center in 2D and 3D domains, respectively:}
    \State \( G_i' \subset G_i \), \( P_j' \subset P_j \) 
    \Comment{Select nearest corners in \(G_i\) and \(P_j\)}

    \State \textbf{2. Calculate minimum distances for each corner in \(P_j'\) to nearest points on \(G_i\):}
    \ForAll{$p \in P_j'$}
        \State $x \gets \arg \min_{x \in G_i} \| p - x \|$
        \State $d_{P\to G} \gets \max_{p \in P_j'} \| p - x \|$
    \EndFor

    \State \textbf{3. Calculate minimum distances for each corner in \(G_i'\) to nearest points on \(P_j\):}
    \ForAll{$g \in G_i'$}
        \State $y \gets \arg \min_{y \in P_j} \| g - y \|$
        \State $d_{G \to P} \gets \max_{g \in G_i'} \| g - y \|$
    \EndFor

    \State \textbf{4. Final contour error between $G_i$ and $P_j$:}
    \State $d_{ij} = \max(d_{P \to G}, d_{G \to P})$
    \State \Return $d_{ij}$
\EndProcedure

\State Initialize distance matrix $D = [d_{ij}]$ of size $(n, m)$
\For{each \(i = 1, \ldots, n\) and \(j = 1, \ldots, m\)}
    \State $D[i, j] \gets \text{Calculate\_Contour\_Error}(G_i, P_j, \text{dim})$
\EndFor

\State Employ Hungarian algorithm to $D$ for optimal assignment

\State \Return $D$

\end{algorithmic}
\end{algorithm}

%\vspace{\baselineskip}  % Adjust value as needed (e.g., -0.5\baselineskip)

All matching criteria, including CE, IoU, GIoU, DIoU, CIoU, CPD, and HD, are computed at each frame of the tracking scenario. At each timestamp, the distance or similarity matrix between the ground truth and predicted boxes is computed, Hungarian matching is employed, and a match whose distance falls within the threshold $\tau$ is counted as a TP for that frame. Unmatched predictions are FPs, and unmatched ground truths are FNs.  A matching criterion, therefore, governs the quality of frame-level detection. 

% \cut{Aggregating these frame-level TPs across the full sequence and across multiple confidence thresholds yields Detection Accuracy (DetA) and the averaged metric mHOTA within the HOTA evaluation protocol~\cite{Luiten_2020}. Association Accuracy (AssA), on the other hand, measures the consistency of the same predicted identity being matched to the same ground truth identity over time. It is computed from the set of frame-level TPs but requires tracking~IDs. Thus, while the matching criterion itself is purely geometric and frame-based, evaluating it within the HOTA framework requires tracker outputs, i.e., predictions that carry persistent identity labels, rather than raw single-frame detections.}

%%%%%%%%%%%%%%%%%%%%%%%%%%%%%%%%%%%%%%%%%%%%%%%%%%%%%%%%%%%%%%%%%%%%%%%%%%%%%%%%
\section{Experimental Evaluation}

We present experiments to demonstrate the capabilities of our proposed Contour Error matching criterion. We adopt the HOTA evaluation protocol~\cite{Luiten_2020} and use the mean Higher-Order Tracking Accuracy (mHOTA) as the primary metric.  At each recall level $r$, HOTA is the geometric mean of Detection Accuracy (DetA) and Association Accuracy (AssA). mHOTA averages HOTA over 40 equally-spaced recall levels: $\text{mHOTA} = \frac{1}{40}\sum_{r}\sqrt{\text{DetA}(r)\cdot\text{AssA}(r)}$. We evaluate seven matching criteria, such as CE, IoU, GIoU, DIoU, CIoU, CPD, and HD, on the nuScenes validation set~\cite{caesar2020nuscenes} using LiDAR-based predictions from the AB3DMOT tracker~\cite{weng2020ab3dmot}. We evaluate the tracker's output, not the raw detections, because the HOTA protocol requires tracking IDs to compute AssA.  Crucially, the tracker is held fixed throughout all experiments. Only the evaluation-stage matching criterion is varied.  This isolates the effect of the matching metric on measured tracking quality, ensuring a fair comparison.

\subsection{Threshold Independent Analysis}\label{sec:threshold_independent}

\begin{figure}[t]
  \centering
  \includegraphics[width=\linewidth]{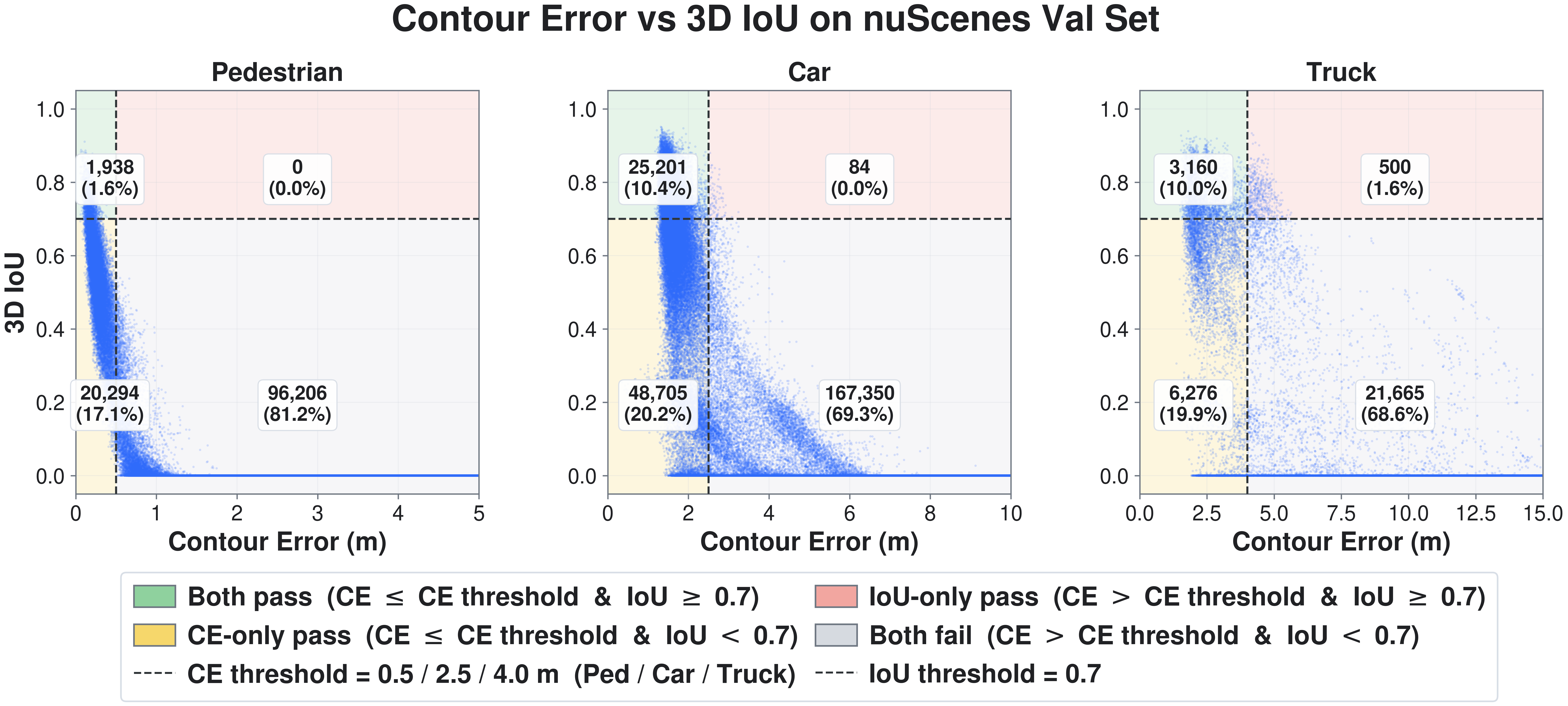}
  \caption{Scatter plots of CE vs.\ IoU for pedestrian, car, and truck categories, within CE distance threshold of \SI{5}{\meter}, \SI{10}{\meter}, and \SI{15}{\meter}, respectively. Background quadrants show both pass (green), CE-only pass (gold), IoU-only pass (coral), and both fail (grey) based on IoU and CE thresholds. Per-quadrant counts annotated.}
  \label{fig:scatter_analysis_iou}
  \vspace{-0.3cm}
\end{figure}

\begin{figure}[t]
  \centering
  \includegraphics[width=\linewidth]{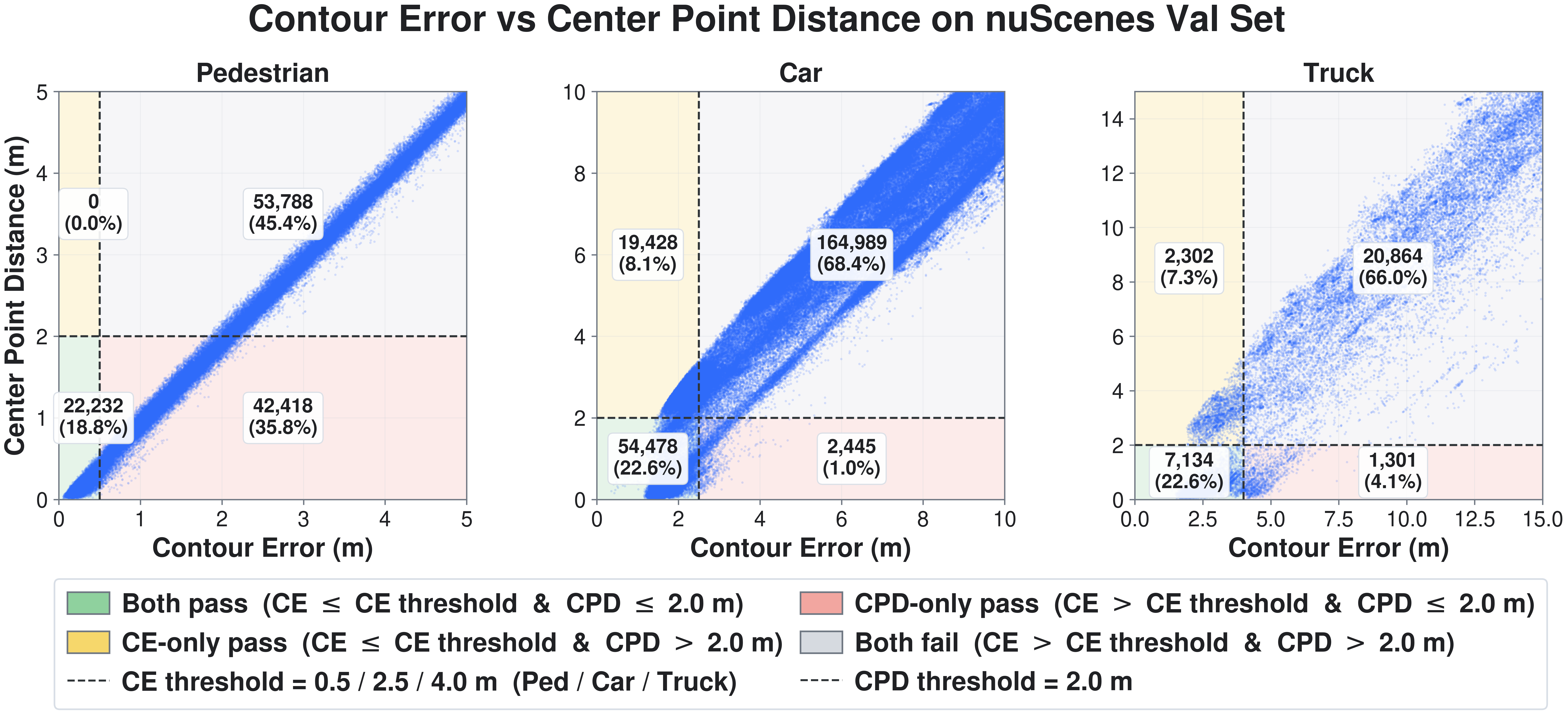}
  \caption{Scatter plots of CE vs.\ CPD for pedestrian, car, and truck categories, within CE distance threshold of \SI{5}{\meter}, \SI{10}{\meter}, and \SI{15}{\meter}, respectively. Background quadrants show both pass (green), CE-only pass (gold), IoU-only pass (coral), and both fail (grey) based on CPD and CE thresholds. Per-quadrant counts annotated.}
    \label{fig:scatter_analysis_cpd}
  \vspace{-0.3cm}
\end{figure}

Figs.~\ref{fig:scatter_analysis_iou} and \ref{fig:scatter_analysis_cpd} present scatter plots of CE vs.\ IoU and CE vs.\ CPD, providing a threshold-independent analysis of the relationship between CE and the standard baseline metrics. Each subplot depicts all pairwise ground truth and prediction combinations within a specified CE range for each category, thereby illustrating the complete distribution without reliance on any specific assignment or operating point. Decision boundaries segment each subplot into four quadrants: IoU~$\geq 0.7$ as defined by the KITTI/Waymo vehicle standard and CPD~$\leq$~\SI{2.0}{\meter} as provided by the nuScenes object detection task. These baseline thresholds are established community settings, not values tuned against CE. CE thresholds are established through sensitivity analysis that maximizes mHOTA, constrained by an upper bound of \SI{50}{\percent} of the BEV diagonal: CE~$\leq$~\SI{0.5}{\meter} for pedestrians, \SI{2.5}{\meter} for cars, and \SI{4.0}{\meter} for trucks. In contrast, Tab.~\ref{tab:match_summary} applies CE-based Hungarian matching (one-to-one assignment) prior to evaluating each matched pair against both CE and the baseline threshold. At IoU~$\geq 0.7$, \SI{75.2}{\percent}, \SI{46.7}{\percent}, and \SI{42.6}{\percent} of CE-matched pairs are CE-only matches for pedestrians, cars, and trucks, respectively, indicating that the ego-facing contour is geometrically close (within $\tau_{\text{CE}}$) despite insufficient volumetric overlap. Fewer than \SI{0.1}{\percent} of car and pedestrian matches satisfy the IoU criterion but fail the CE criterion, demonstrating that CE encompasses nearly all IoU-valid matches while also identifying a substantial additional set that IoU excludes.

% Table I (correlations) omitted for brevity; correlations are visible directly from the scatter plots in Figs.~\ref{fig:scatter_analysis_iou}-\ref{fig:scatter_analysis_cpd}.
% \begin{table}
% \centering
% \footnotesize
% \caption{Pairwise correlations among CE, IoU, and CPD across nuScenes object categories.}
% \label{tab:correlation_summary}
% \begin{tabular}{P{1.25cm}|P{1.75cm}|P{1.75cm}|P{1.75cm}}
% \toprule
% \textbf{Object Category} & \textbf{Corr(Contour, IoU)} & \textbf{Corr(Contour, CPD)} & \textbf{Corr(IoU, CPD)} \\ 
% \midrule
% Pedestrian & $-0.75$ & $+0.775$ & $-0.79$ \\
% Car        & $-0.71$ & $+0.752$ & $-0.96$ \\
% Truck      & $-0.44$ & $+0.536$ & $-0.94$ \\
% \bottomrule
% \end{tabular}
% \vspace{-0.2cm}
% \end{table}

\begin{figure*}[t] 
\centering 
\includegraphics[width=0.95\linewidth]{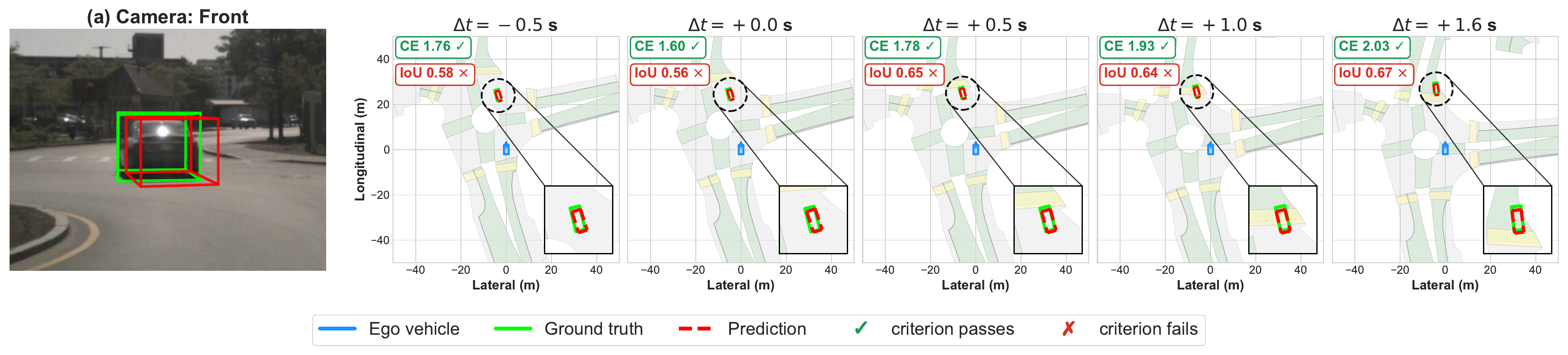} 
\vspace{2pt} 
\includegraphics[width=0.95\linewidth]{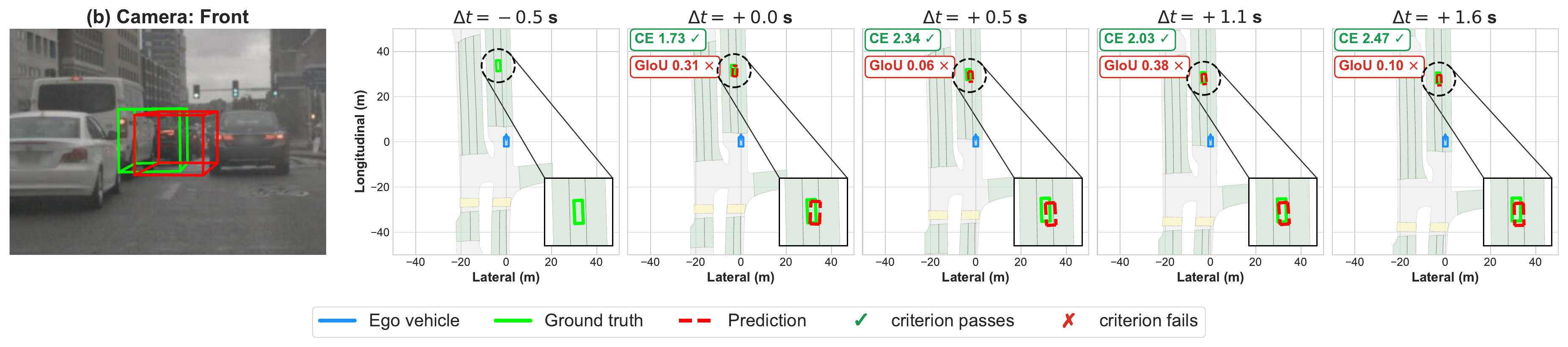}
\caption{Temporal failure cases in the nuScenes Val set: two in-lane objects (top and bottom) for which CE matches across all five frames, but the overlap-based
criterion persistently fails. Each row shows the camera view (left), five BEV frames with per-frame metric values (green = pass, red = fail), and a zoomed inset (black outline) that magnifies the ego-facing object so the GT-prediction geometry is visible. Ground-truth, predictions, and ego boxes are represented in green, red, and blue, respectively. (a) CE vs. IoU~$\geq 0.7$ with object at \SI{25}{\meter}. (b) CE vs. GIoU~$\geq 0.5$ with object at \SI{31.8}{\meter} in rain.}
\label{fig:car_truck_qualitative} 
\vspace{-0.3cm}
\end{figure*}

\begin{table}[t]
\centering
\caption{Quadrant Distribution of Unique CE-Based  Hungarian Matches against defined thresholds on nuScenes val set.}
% Cells give the share (\si{\percent}) of unique matches per pass/fail quadrant against IoU ($\geq 0.7$) and CPD ($\leq\SI{2.0}{\meter}$). CE Thresholds: \SI{0.5}{}/\SI{2.5}{}/\SI{4.0}{\meter}}
\label{tab:match_summary}
\resizebox{\columnwidth}{!}{%
\begin{tabular}{@{}lr|rrrr|rrrr@{}}
\toprule
& & \multicolumn{4}{c|}{\emph{CE vs.\ IoU ($\geq 0.7$)}} & \multicolumn{4}{c}{\emph{CE vs.\ CPD ($\leq \SI{2.0}{\meter}$)}} \\
\cmidrule(lr){3-6} \cmidrule(l){7-10}
\textbf{Class} & \makecell{\textbf{Unique}\\\textbf{matches}} & \makecell{\textbf{Both}\\\textbf{pass}} & \makecell{\textbf{CE}\\\textbf{only}} & \makecell{\textbf{Both}\\\textbf{fail}} & \makecell{\textbf{IoU}\\\textbf{only}} & \makecell{\textbf{Both}\\\textbf{pass}} & \makecell{\textbf{CE}\\\textbf{only}} & \makecell{\textbf{Both}\\\textbf{fail}} & \makecell{\textbf{CPD}\\\textbf{only}} \\
\midrule
Ped.  & \num{25224} & \num{7.7} & \num{75.2} & \num{17.2} & \num{0.0} & \num{82.8} &  \num{0.0} &  \num{7.0} & \num{10.2} \\
Car   & \num{58285} & \num{42.9} & \num{46.7} & \num{10.3} & \num{0.1} & \num{87.7} &  \num{1.9} &  \num{7.7} &  \num{2.7} \\
Truck &  \num{9650} & \num{32.3} & \num{42.6} & \num{20.0} & \num{5.1} & \num{70.4} &  \num{4.5} & \num{12.8} & \num{12.3} \\
\bottomrule
\end{tabular}%
}
%\vspace{-0.2cm}
\end{table}

\begin{table}[t]
\centering
\footnotesize
\caption{Matching Criteria by Ego Distance for Car Category on nuScenes val set (Recall@0.5). \textbf{Bold}/\underline{Underline}: Best/Second-best count.}
\label{tab:car_proximity_summary}
\setlength{\tabcolsep}{2pt}
\renewcommand{\arraystretch}{0.95}
\begin{tabular}{l*{6}{r}}
\toprule
\multirow{2}{*}{\makecell{\textbf{Matching}\\\textbf{Criterion}}} &
\multicolumn{3}{c}{\textbf{TPs} $\uparrow$} &
\multicolumn{3}{c}{\textbf{Failures (FNs)} $\downarrow$}\\
\cmidrule(lr){2-4}\cmidrule(lr){5-7}
& \makecell{0-\SI{10}{\meter}}
& \makecell{10-\SI{20}{\meter}}
& \makecell{20-\SI{30}{\meter}}
& \makecell{0-\SI{10}{\meter}}
& \makecell{10-\SI{20}{\meter}}
& \makecell{20-\SI{30}{\meter}}\\
\midrule
\multicolumn{7}{l}{\emph{Overlap-based (object-centric)}} \\
3D IoU  & \num{4960} & \num{8544} & \num{6783} & \num{581} & \num{3832} & \num{8260} \\
GIoU    & \num{4936} & \num{8457} & \num{6634} & \num{605} & \num{3919} & \num{8409} \\
DIoU / CIoU & \num{4950} & \num{8520} & \num{6760} & \num{591} & \num{3856} & \num{8283} \\
\midrule
\multicolumn{7}{l}{\emph{Distance-based (object-centric)}} \\
CPD   & \textbf{\num[detect-weight=true]{4970}} & \textbf{\num[detect-weight=true]{8695}} & \textbf{\num[detect-weight=true]{7069}} & \textbf{\num[detect-weight=true]{571}} & \textbf{\num[detect-weight=true]{3681}} & \textbf{\num[detect-weight=true]{7974}} \\
HD    & \num{4960} & \num{8654} & \num{7026} & \num{581} & \num{3722} & \num{8017} \\
\midrule
\rowcolor{rowlabelcolor}\multicolumn{7}{l}{\emph{Distance-based (ego-centric - Ours)}} \\
\rowcolor{rowlabelcolor} CE ($k\!=\!6$) & \underline{\num{4962}} & \underline{\num{8655}} & \underline{\num{7029}} & \underline{\num{579}} & \underline{\num{3721}} & \underline{\num{8014}} \\
\bottomrule
\end{tabular}
\vspace{-0.3cm}
\end{table}

\begin{table}[t]
\centering
\footnotesize
\caption{Matching Criteria by Yaw-Error Severity within \SI{30}{\meter} for Car Category on nuScenes val set (Recall@0.5). \textbf{Bold}/\underline{Underline}: Best/Second-best count.}
\label{tab:car_yaw_angle_summary}
\setlength{\tabcolsep}{2pt}
\renewcommand{\arraystretch}{0.95}
\begin{tabular}{l*{6}{r}}
\toprule
\multirow{2}{*}{\makecell{\textbf{Matching}\\\textbf{Criterion}}} &
\multicolumn{3}{c}{\textbf{TPs} $\uparrow$} &
\multicolumn{3}{c}{\textbf{Failures (FNs)} $\downarrow$}\\
\cmidrule(lr){2-4}\cmidrule(lr){5-7}
& \makecell{\textbf{L}\\($<\!10^\circ$)}
& \makecell{\textbf{M}\\(10-30$^\circ$)}
& \makecell{\textbf{H}\\($>\!30^\circ$)}
& \makecell{\textbf{L}\\($<\!10^\circ$)}
& \makecell{\textbf{M}\\(10-30$^\circ$)}
& \makecell{\textbf{H}\\($>\!30^\circ$)}\\
\midrule
\multicolumn{7}{l}{\emph{Overlap-based (object-centric)}} \\
3D IoU  & \num{2513} & \num{388} & \num{1424} & \num{26} & \num{13} & \num{50} \\
GIoU    & \num{2500} & \num{382} & \num{1411} & \num{39} & \num{19} & \num{63} \\
DIoU / CIoU & \num{2509} & \num{388} & \num{1422} & \num{30} & \num{13} & \num{52} \\
\midrule
\multicolumn{7}{l}{\emph{Distance-based (object-centric)}} \\
CPD   & \textbf{\num[detect-weight=true]{2538}} & \textbf{\num[detect-weight=true]{401}} & \textbf{\num[detect-weight=true]{1467}} & \textbf{\num[detect-weight=true]{1}} & \textbf{\num[detect-weight=true]{0}} & \textbf{\num[detect-weight=true]{7}} \\
HD    & \num{2528} & \num{400} & \num{1462} & \num{11} & \num{1} & \num{12} \\
\midrule
\rowcolor{rowlabelcolor}\multicolumn{7}{l}{\emph{Distance-based (ego-centric - Ours)}} \\
\rowcolor{rowlabelcolor} CE ($k\!=\!6$) & \underline{\num{2529}} & \underline{\num{400}} & \underline{\num{1463}} & \underline{\num{10}} & \underline{\num{1}} & \underline{\num{11}} \\
\bottomrule
\end{tabular}
\vspace{-0.3cm}
\end{table}

\subsection{Proximity and Yaw Error Analysis}
To compare all seven matching criteria under controlled conditions, we evaluate every GT box across all 150 validation scenes at a fixed operating point, Recall@0.5, corresponding to a confidence score threshold of 0.635 for cars. This threshold is calibrated using CE-based matching, and the same confidence threshold is then applied identically to all seven criteria, ensuring a common set of active predictions. For each criterion at its standard threshold, we perform per-frame Hungarian matching and classify each GT box as a true positive (TP) or a failure (FN), i.e., a GT box that remains unmatched (FN) under the criterion's threshold. We then bin the results by the ego to GT distance and yaw angle error as shown in Tab.~\ref{tab:car_proximity_summary} and Tab.~\ref{tab:car_yaw_angle_summary}, respectively.

\paragraph{Proximity-Based Analysis}
We partition the ego to GT distance into three bins \SIrange{0}{10}{\meter}, \SIrange{10}{20}{\meter}, and \SIrange{20}{30}{\meter}, reflecting the safety-critical ranges where closer objects demand more reliable detection as illustrated in Tab.~\ref{tab:car_proximity_summary}. Across all three proximity bins, the distance-based criteria (CPD, HD, CE) consistently pass more GT-prediction pairs through the matching gate than the overlap-based criteria (IoU, GIoU, DIoU/CIoU). This gap widens with range in the closest bin from \SIrange{0}{10}{\meter}, where the difference between the most permissive CPD and most restrictive GIoU is only 34 TPs, but in the \SIrange{20}{30}{\meter} bin it grows to 435 TPs. Among the overlap family, the standard IoU yields more matches than its generalized variants. Among the distance-based criteria, CPD is the most permissive, with CE and HD within 1\% of CPD across all bins. CE yields the second-lowest failure count in every bin, indicating that its ego-centric weighting preserves the permissiveness of distance-based matching while still penalizing translational, shape, and orientation errors, a property that positions it between the overly strict IoU and the orientation-blind CPD.

\paragraph{Yaw Error Analysis} To isolate the effect of orientation misalignment, Tab.~\ref{tab:car_yaw_angle_summary} bins GT boxes within \SI{30}{\meter} by the yaw error between each GT and its matched prediction. We compute the yaw error using the prediction that matches the union of all criteria, i.e., a GT box is included if any criterion matches it, ensuring a common reference set. CPD, being purely translational, is nearly immune to yaw error (\num{8} total failures), but this immunity is precisely its limitation as a safety metric. It certifies predictions as correct regardless of heading error, but provides no assurance about the predicted trajectory direction. The overlap-based criteria sit at the opposite extreme, accumulating \num{50}-\num{63} failures at high severity ($>$\,\num{30}\textdegree) as even moderate rotations rapidly destroy box overlap. CE and HD occupy a principled middle ground with \num{11}-\num{12} high-severity failures. In terms of failure rate defined as failures normalized by total GT boxes per bin, CE yields \num{10}/\num{2539}\,=\,\num{0.39}\% at low severity, \num{1}/\num{401}\,=\,\num{0.25}\% at medium, and \num{11}/\num{1474}\,=\,\num{0.75}\% at high. The medium-bin rate is not higher than the low-bin rate, but that bin contains only a single failure, so the estimate is dominated by sampling uncertainty; the high-severity rate is clearly elevated relative to low severity. Absolute counts are also influenced by unequal bin sizes. For car-sized boxes ($4.6 \times \SI{1.9}{\meter}$), a $30^\circ$ yaw shift displaces the nearest ego-facing corner by $\approx$\,\SI{1}{\meter}, well within CE's $\SI{2.5}{\meter}$ threshold. Thus,  CE appropriately tolerates heading errors that do not significantly alter the ego-facing contour, whereas the overlap criteria over-penalize them.

% \begin{figure}[t]
% \centering
% \begin{tabular}{@{}c@{\hspace{4pt}}c@{}}
% \includegraphics[width=0.48\columnwidth]{figures/threshold_saturation.png}
% \label{fig:mHOTA} 
% &
% \includegraphics[width=0.48\columnwidth]{figures/corner_ablation_extended.png}
% \label{fig:corner_ablation}\\
% \footnotesize (a) mHOTA threshold saturation & \footnotesize (b) Corner ablation ($k$) \\  
% \end{tabular}
% \caption{(a) mHOTA vs.\ normalized matching threshold for all seven criteria (DIoU/CIoU merged) on the nuScenes val set (Car, $\tau{=}\SI{2.5}{\meter}$). CE captures 98.9\% of its peak mHOTA at $\tau{=}\SI{2.5}{\meter}$ and fully saturates at $\tau \geq \SI{3.5}{\meter}$. (b) DetA and AssA vs.\ the number of ego-centric corners $k$ at $\tau{=}\SI{2.5}{\meter}$. Varying $k$ only affects frame-level matching (DetA), not track-level identity (AssA). DetA varies by $<$1.3\%, confirming robustness. We select $k{=}6$.}
% \label{fig:saturation_ablation}
% \vspace{-0.3cm}
% \end{figure}

\subsection{Temporal Failure Case Analysis}\label{sec:temporal_analysis}

To demonstrate the practical consequences of disagreements between matching criteria, the full nuScenes validation set was analyzed for instances where CE ($k{=}6$) successfully matched a ground truth (GT) and prediction pair, but the competing overlap-based metric did not. Candidates were restricted to in-lane objects located ahead of the ego vehicle and visible in the front camera, and were tracked over a five-frame window (\SI{2}{\second}). Figure~\ref{fig:car_truck_qualitative} presents two representative examples. (a) IoU ($\geq 0.7$) fails in all five frames for a car at \SI{25}{\meter}, with values ranging from \num{0.56} to \num{0.67}, while CE passes throughout with values between \num{1.60} and \num{2.03}\,\si{\meter}. This outcome arises because IoU computes volumetric intersection, which is highly sensitive to minor translational and rotational offsets. A slight shift along the box's long axis can significantly reduce the intersection volume, while having minimal impact on the ego-facing contour distance. (b) GIoU ($\geq 0.5$) fails in every frame where the prediction exists for a car at \SI{31.8}{\meter} in rain, with values ranging from \num{0.06} to \num{0.38}, while CE passes with values between \num{1.73} and \num{2.47}\,\si{\meter}. GIoU introduces a penalty based on the smallest enclosing box, which increases rapidly when the prediction is offset or rotated, causing GIoU to fall below its threshold even when the ego-facing surfaces remain close. In both scenarios, CE's contour-based formulation measures only the maximum surface deviation of the $k$ nearest corners, making it inherently robust to the far-side geometry that dominates IoU and GIoU. The overlap-based rejection would cause the tracker to lose identity continuity for an object within the ego vehicle's driving corridor, resulting in a persistent tracking-level failure across the entire temporal window.

\begin{table}[t]
\centering
\footnotesize
\caption{Aggregate Tracking for Car Category on nuScenes val set. mHOTA/AMOTA averaged over \num{40} recall levels; remaining columns at Recall@0.5. \textbf{Bold}/\underline{Underline}: Best/Second-best.}
\label{tab:main-optimal}
\setlength{\tabcolsep}{2pt}
\renewcommand{\arraystretch}{0.95}
\begin{tabular}{l*{7}{r}}
\toprule
\makecell{\textbf{Matching}\\\textbf{Criterion}} &
\makecell{\textbf{mHOTA}\\$\uparrow$} &
\makecell{\textbf{AMOTA}\\$\uparrow$} &
\makecell{\textbf{AssA}\\$\uparrow$} &
\makecell{\textbf{DetA}\\$\uparrow$} &
\makecell{\textbf{TPs}\\$\uparrow$} &
\makecell{\textbf{FPs}\\$\downarrow$} &
\makecell{\textbf{FNs}\\$\downarrow$}\\
\midrule
\multicolumn{8}{l}{\emph{Overlap-based (object-centric)}} \\
3D IoU  & \num{0.633} & \num{0.470} & \textbf{\num[detect-weight=true]{0.991}} & \num{0.431} & \num{26214} & \num{2499} & \num{32103} \\
GIoU    & \num{0.618} & \num{0.454} & \num{0.991} & \num{0.416} & \num{25572} & \num{3141} & \num{32745} \\
DIoU / CIoU & \num{0.630} & \num{0.468} & \num{0.991} & \num{0.429} & \num{26107} & \num{2606} & \num{32210} \\
\midrule
\multicolumn{8}{l}{\emph{Distance-based (object-centric)}} \\
CPD   & \textbf{\num[detect-weight=true]{0.664}} & \textbf{\num[detect-weight=true]{0.505}} & \num{0.989} & \textbf{\num[detect-weight=true]{0.463}} & \textbf{\num[detect-weight=true]{27534}} & \textbf{\num[detect-weight=true]{1179}} & \textbf{\num[detect-weight=true]{30783}} \\
HD    & \num{0.654} & \num{0.499} & \num{0.982} & \num{0.457} & \num{27289} & \num{1424} & \num{31028} \\
\midrule
\rowcolor{rowlabelcolor}\multicolumn{8}{l}{\emph{Distance-based (ego-centric - Ours)}} \\
\rowcolor{rowlabelcolor} CE ($k\!=\!6$) & \underline{\num{0.654}} & \underline{\num{0.499}} & \underline{\num{0.982}} & \underline{\num{0.457}} & \underline{\num{27307}} & \underline{\num{1406}} & \underline{\num{31010}} \\
\bottomrule
\end{tabular}
\vspace{-0.3cm}
\end{table}

\subsection{Aggregate Tracking Performance}\label{sec:beyond_mhota}

Table~\ref{tab:main-optimal} presents a comparison of all matching criteria at their standard thresholds (IoU family $\geq 0.5$, CPD $\leq\SI{2.0}{\meter}$, CE/HD $\leq\SI{2.5}{\meter}$). The aggregate HOTA comparison uses the established $\geq 0.5$ IoU-family setting, whereas the stricter $\geq 0.7$ IoU setting is used for the KITTI/Waymo comparison in the scatter analysis; neither is tuned against CE. CPD attains the highest mHOTA (\num{0.664}) because it reduces each box to its center point, thereby omitting information about shape or orientation. Notably, a $90^\circ$ incorrect heading results in CPD\,=\,0. The higher true positive (TP) count for CPD reflects a more permissive matching criterion rather than a more accurate evaluation. CE ($k{=}6$) at $\tau{=}\SI{2.5}{\meter}$ achieves mHOTA\,=\,0.654, surpassing all IoU variants ($\geq 0.5$) by 2.1 to 3.6 percentage points. This gap increases at the stricter threshold of $\geq 0.7$ used by KITTI and Waymo benchmarks. While HD achieves the same aggregate mHOTA, CE consistently identifies slightly more TPs (\num{27307} vs.\ \num{27289}). The ego-centric restriction in CE excludes far-side corners, whose estimation noise would otherwise inflate the Hausdorff distance without contributing to safety. A TP under CE guarantees that the ego-facing surface is within $\tau$ of the ground truth (GT), a property not ensured by either HD or CPD.

A key observation is that the choice of matching criterion significantly affects frame-level DetA ($\Delta{=}0.047$), whereas AssA remains nearly unchanged ($\Delta{<}0.01$). This finding indicates that the matching criterion determines which pairs are classified as correct detections, while the tracker's identity assignment is largely unaffected. DIoU and CIoU yield identical match assignments because CIoU's aspect-ratio penalty $v$ is approximately zero for trained detectors~\cite{Zheng2019DistanceIoULF}. Therefore, these criteria are merged in subsequent analysis. However, aggregate mHOTA does not fully capture safety-critical behavior. Temporal failures illustrated in Fig.~\ref{fig:car_truck_qualitative} demonstrate that overlap-based criteria reject detections with close contour proximity, resulting in persistent identity loss, whereas CPD's translational formulation accepts heading-flipped predictions without penalty. Consequently, aggregate rankings should not serve as the sole basis for selecting a matching criterion. The geometric interpretation of a true positive must also be considered.

\begin{figure}[t]
\centering
\begin{subfigure}[b]{0.48\columnwidth}
  \centering
  \includegraphics[width=\linewidth]{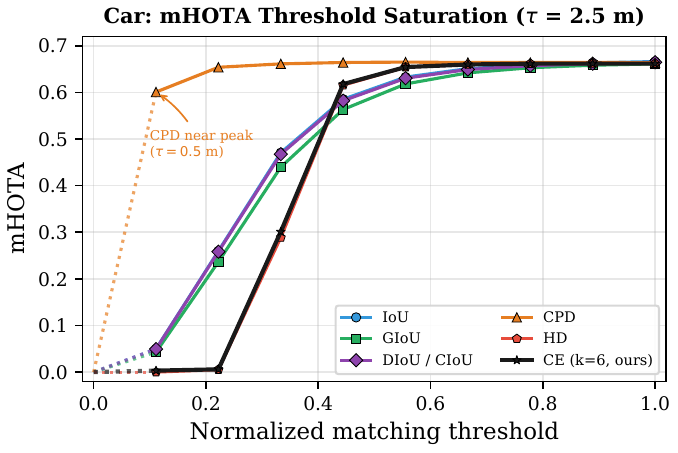}
  \caption{mHOTA threshold saturation}
  \label{fig:mHOTA}
\end{subfigure}
\hfill
\begin{subfigure}[b]{0.48\columnwidth}
  \centering
  \includegraphics[width=\linewidth]{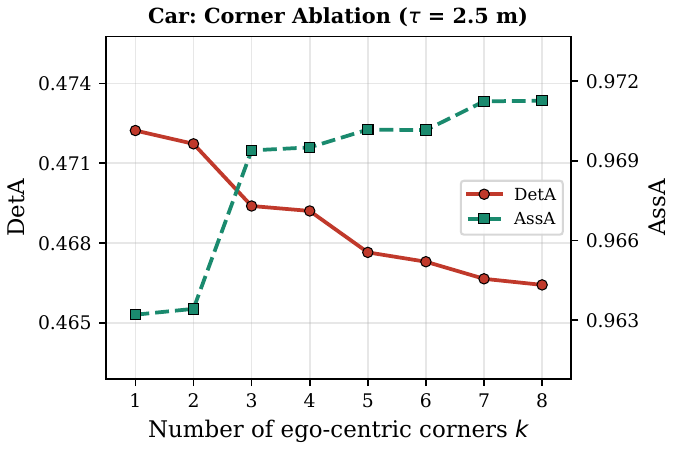}
  \caption{Corner ablation ($k$)}
  \label{fig:corner_ablation}
\end{subfigure}
\caption{(a) mHOTA vs.\ normalized matching threshold for all matching criteria on the nuScenes val set (Car, $\tau{=}\SI{2.5}{\meter}$). CE captures \SI{98.9}{\percent} of its peak mHOTA at $\tau{=}\SI{2.5}{\meter}$ and fully saturates at $\tau \geq \SI{3.5}{\meter}$. (b) DetA and AssA vs.\ $k$ at $\tau{=}\SI{2.5}{\meter}$ (shared absolute axis span). Both vary by less than one percentage point across $k$.}
\label{fig:saturation_ablation}
\vspace{-0.5cm}
\end{figure}

\subsection{Threshold Saturation and Corner Ablation}\label{sec:saturation_ablation}

Three key design questions are addressed: (i) sensitivity to the distance threshold $\tau$, (ii) sensitivity to the number of corners $k$, and (iii) the relationship between CE and HD. Figure~\ref{fig:saturation_ablation} presents a comprehensive analysis across nine thresholds (\SIrange{0.5}{5.0}{\meter}) and $k\in\{1,\ldots,8\}$ corners. Figure~\ref{fig:mHOTA} illustrates how mHOTA varies with a normalized matching threshold for all six criteria. By construction mHOTA$\to 0$ as $\tau\to 0$; the leftmost markers (normalized $\approx 0.1$) correspond to $\tau{=}\SI{0.5}{\meter}$, where CPD is already near peak because only centers must agree, whereas CE and HD remain near zero until $\tau$ grows. CE achieves \SI{98.9}{\percent} of its peak mHOTA at $\tau = \SI{2.5}{\meter}$ and fully saturates at $\tau \geq \SI{3.5}{\meter}$. This result confirms that the recommended threshold captures nearly all of CE's discriminative power while remaining within a geometrically meaningful range (approximately 50\% of the car BEV diagonal). IoU variants saturate more slowly because the overlap criterion is inherently stricter.

Figure~\ref{fig:corner_ablation} presents the corner ablation at $\tau = \SI{2.5}{\meter}$. Varying $k$ alters the classification of ground truth (GT) and prediction pairs as TPs, FPs, or FNs at each frame. Both Detection Accuracy (DetA) and Association Accuracy (AssA) respond as the shifts are small and of comparable absolute magnitude. The AssA drifts because it is computed over the changing TP set rather than from a change in tracker-identity persistence. DetA and AssA vary by less than \num{1.3}\% across all $k$ values, confirming strong robustness. The choice $k = 6$ is motivated by typical driving geometry. In the most common scenario, where a vehicle is ahead and slightly offset, the ego sensor observes two faces of the cuboid (the rear and one side), thereby exposing exactly six unique vertices. The remaining two corners on the far side are occluded and carry the highest estimation uncertainty. When only one face is visible, as with a perfectly parallel vehicle, four corners are observable, and the $k{=}6$ selection conservatively includes two additional near-side and far-side corners. The 2D counterpart $k{=}3$ follows the same principle, as three corners spanning two adjacent edges are required to jointly capture translational and yaw error. At $k = 8$, CE becomes the object-centric Hausdorff-distance criterion and thus, the $k{=}8$ point in Fig.~\ref{fig:corner_ablation} provides the object-centric comparison.

%%%%%%%%%%%%%%%%%%%%%%%%%%%%%%%%%%%%%%%%%%%%%%%%%%%%%%%%%%%%%%%%%%%%%%%%%%%%%%%%
\section{Discussion}\label{sec:discussion}

\paragraph{Influence of threshold selection}
Each matching criterion requires a threshold to partition ground truth and prediction pairs into TPs and FNs. This selection directly influences the measured tracking performance. Fig.~\ref{fig:mHOTA} demonstrates that CE's mHOTA achieves 98.9\% of its peak at $\tau{=}\SI{2.5}{\meter}$ and saturates beyond $\tau{\geq}\SI{3.5}{\meter}$. This indicates low sensitivity within the practical operating range. However, no single threshold can be considered objectively correct. To address threshold dependence, one approach is to integrate performance over a range of thresholds. This is similar to the COCO~\cite{lin2014microsoft} methodology, which averages average precision (AP) over IoU thresholds from 0.5 to 0.95. Calculating a threshold-integrated mHOTA by averaging across a standardized range of CE thresholds (e.g., \SIrange{0.5}{5.0}{\meter}) would enable a more comprehensive comparison than any single operating point. This represents an important direction for future research.

\paragraph{Distance-dependent and criticality-aware thresholds}
A fixed threshold applies the same \SI{2.5}{\meter} contour error regardless of whether the object is \SI{5}{\meter} or \SI{50}{\meter} away. In practice, perception uncertainty increases with distance. The safety relevance of a given geometric error decreases with range. Distance-dependent thresholds are stricter at close range (e.g., $\tau{=}\SI{1.0}{\meter}$ within \SI{10}{\meter}) and more relaxed at greater distances (e.g., $\tau{=}\SI{4.0}{\meter}$ beyond \SI{30}{\meter}). Such thresholds can encode this prior knowledge. In terms of azimuth angle, a constant angular threshold naturally yields tighter absolute tolerances for nearby objects and looser tolerances for distant objects. This partially reflects ego-centric criticality without requiring an explicit planner. These distance-dependent criteria bridge the gap between purely geometric matching and criticality-aware evaluation. They provide a practical compromise for open-loop testing when a planner is unavailable.

\begin{table}[t]
\centering
\footnotesize
\caption{Per-Pair Matching Cost over nuScenes GT-Pred Pairs. Theoretical complexity and measured wall-clock time. $k$: corners, $n_e$: edges, $n_v$: vertices, $s$: surface samples.}
\label{tab:computational_cost}
\begin{tabular}{llr}
\toprule
\textbf{Criterion} & \textbf{Complexity} & \textbf{$\mu$s/pair} \\
\midrule
\multicolumn{3}{l}{\emph{Object-centric metrics}} \\
CPD & $\mathcal{O}(1)$ & \num{6} \\
3D IoU & $\mathcal{O}(n_v^2)$ & \num{377} \\
HD / CE $k{=}8$ (3D) & $\mathcal{O}(8 \cdot n_e)$ & \num{544} \\
GIoU / DIoU / CIoU & $\mathcal{O}(n_v^2)$ & \num{549}-\num{644} \\
Hausdorff (cont., $s{=}100$) & $\mathcal{O}(s^2)$ & \num{2887} \\
\midrule
\rowcolor{rowlabelcolor}\multicolumn{3}{l}{\emph{Ego-centric metrics (Ours)}} \\
\rowcolor{rowlabelcolor} CE $k{=}3$ (2D) & $\mathcal{O}(k \cdot n_e)$ & \num{380} \\
\rowcolor{rowlabelcolor} CE $k{=}6$ (3D) & $\mathcal{O}(k \cdot n_e)$ & \num{592} \\
\bottomrule
\end{tabular}
\vspace{-0.3cm}
\end{table}

\paragraph{Relationship to downstream evaluation}
Matching is a single component within the open-loop evaluation pipeline. Establishing a connection between matching quality and downstream safety remains an unresolved challenge. One example is time-to-collision (TTC) conditioning or planner-centric metrics~\cite{philion2020learning, ivanovic2022injecting}. The ego-centric design of CE prioritizes the surface geometry most relevant to collision checking. However, comprehensive closed-loop validation is beyond the scope of this work.

\paragraph{Difficulty of an objective meta-criterion}
A higher mHOTA score does not always indicate a superior criterion. It may simply reflect a more permissive threshold. Ideally, true positives should correspond to application-relevant correctness. False negatives and false positives should be aligned with critical driving behaviors. Rigorous establishment of this relationship would require either closed-loop simulation, which validates matching decisions against planner outputs, or human annotation by experts who manually match predictions to ground truth. The latter is challenging because annotators disagree across backgrounds and protocols. Future research could conduct controlled annotation studies to provide an empirical foundation for meta-evaluation.

\paragraph{Computational cost}
Table~\ref{tab:computational_cost} presents both theoretical complexity and measured per-pair wall-clock times for \num{2000} nuScenes ground truth and prediction pairs. All discrete-corner criteria operate within the same order of magnitude (377–644\,\si{\micro\second}). CE ($k{=}6$) is marginally slower than object-centric Hausdorff distance (HD, $k{=}8$). HD corresponds to the object-centric limit of CE with all eight box corners ($k{=}8$); CE's small runtime overhead arises from sorting corners by ego distance and selecting the $k$ nearest corners. This is despite CE evaluating fewer corners. CE requires an additional step to sort by ego-distance and select the $k$ nearest corners. This overhead is absent in HD, which uses all corners without selection. The resulting ${\approx}$9\% overhead is the computational cost of ego-centric awareness. At \num{592}\,\si{\micro\second} per GT-prediction pair, CE remains in the same computational order as 3D IoU and is negligible relative to a perception cycle. All discrete methods are substantially faster than continuous Hausdorff ($s{=}100$: 2,887\,\si{\micro\second}, approximately five times slower).

%%%%%%%%%%%%%%%%%%%%%%%%%%%%%%%%%%%%%%%%%%%%%%%%%%%%%%%%%%%%%%%%%%%%%%%%%%%%%%%%
\section{Conclusion}

We introduced Contour Errors (CE) as an ego-centric matching criterion for open-loop evaluation of 3D multi-object tracking in autonomous driving. CE employs Hausdorff-type reasoning to sparse bounding-box corner geometry, utilizing $k$-nearest ego-centric corner selection. This approach offers graded orientation sensitivity between the extremes of Intersection over Union (IoU), which over-penalizes minor yaw deviations, and Closest Point Distance (CPD), which is insensitive to orientation. CE results in only \num{11} high-yaw failures, compared to \num{50}–\num{63} for the IoU variants, while CPD's \num{8} failures demonstrate its insensitivity to heading error. Extensive experiments on nuScenes across seven matching criteria indicate that CE outperforms all overlap-based criteria by \num{2}–\num{4} percentage points in mHOTA. Future research directions include threshold-integrated mHOTA, which averages over a standardized range of thresholds, analogous to COCO Average Precision (AP), distance-dependent thresholds that encode range-varying criticality, and the use of CE as a differentiable training loss to optimize detectors for ego-centric contour alignment.

%%%%%%%%%%%%%%%%%%%%%%%%%%%%%%%%%%%%%%%%%%%%%%%%%%%%%%%%%%%%%%%%%%%%%%%%%%%%%%%%

% Bibliography setup
% \bibliographystyle{ieeetr} % IEEE style for BibTeX
%\printbibliography
\footnotesize
\bibliographystyle{IEEEtran}
\bibliography{references.bib}

\end{document}